\documentclass[conference]{IEEEtran}
\IEEEoverridecommandlockouts
\usepackage{cite}
\usepackage{amsmath,amssymb,amsfonts}
\usepackage{algorithmic}
\usepackage{graphicx}
\usepackage{graphics}
\usepackage{textcomp}
\usepackage{pgfplots}
\usepackage{xcolor}
\usepackage{colortbl}
\usepackage{booktabs} 
\usepackage{subfig} 
\usepackage{multirow} 
\usepackage{hyperref} 
\usepackage{todonotes} 
\makeatletter 
\newcommand{\linebreakand}{%
  \end{@IEEEauthorhalign}
  \hfill\mbox{}\par
  \mbox{}\hfill\begin{@IEEEauthorhalign}
}
\makeatother 

\def\BibTeX{{\rm B\kern-.05em{\sc i\kern-.025em b}\kern-.08em
    T\kern-.1667em\lower.7ex\hbox{E}\kern-.125emX}}
\begin{document}

\title{ Domain Generalization for Endoscopic Image Segmentation by
Disentangling Style-Content Information and SuperPixel Consistency}

\author{ \IEEEauthorblockN{1\textsuperscript{st} Mansoor Ali Teevno}

\IEEEauthorblockA{\textit{School of Engineering and Sciences } \\
\textit{Tecnologico de Monterrey}\\
Monterrey, Mexico \\
Email: a01753093@tec.mx}
\and
\IEEEauthorblockN{2\textsuperscript{nd} Rafael Martinez-Garcia-Peña}

\IEEEauthorblockA{\textit{School of Engineering and Sciences } \\
\textit{Tecnologico de Monterrey}\\
Monterrey, Mexico \\
Email: rafael.mgp@tec.mx}

\and
\IEEEauthorblockN{3\textsuperscript{rd} Gilberto Ochoa-Ruiz}

\IEEEauthorblockA{\textit{School of Engineering and Sciences } \\
\textit{Tecnologico de Monterrey}\\
Monterrey, Mexico \\
Email: gilberto.ochoa@tec.mx }
\linebreakand 
\IEEEauthorblockN{4\textsuperscript{th} Sharib Ali}

\IEEEauthorblockA{\textit{School of Computing } \\
\textit{University of Leeds}\\
Leeds, UK \\
Email: S.S.Ali@leeds.ac.uk }}


\maketitle

\begin{abstract}
Frequent monitoring is necessary to stratify individuals based on their likelihood of developing gastrointestinal (GI) cancer precursors. In the clinical practice, white-light imaging (WLI), and complimentary modalities such as narrow-band imaging (NBI) and fluorescence imaging are used to assess risk areas. However, conventional deep learning (DL) models have depleted performance due to domain gap when a model is trained on one modality and tested on a different one. In our earlier approach we used superpixel based method referred to as ``SUPRA'' to effectively learn domain-invariant information using color and space distances to generate groups of pixels. One of the main limitations of this early work is that the aggregation does not exploit structural information, making it sub-optimal for segmentation tasks, especially for polyps and heterogeneous color distributions. Therefore, in this work, we propose an approach for style-content disentanglement using instance normalization and instance selective whitening (ISW) for an improved domain generalization when combined with SUPRA. We evaluate our approach on two datasets: EndoUDA Barret’s Esophagus and EndoUDA polyps and compare its performance with previous three state-of-the-art (SOTA) methods. Our findings demonstrate a notable enhancement in performance compared to both baseline and state-of-the-art methods across the target domain data. Specifically, our approach exhibited improvements of 14\%, 10\%, 8\%, and 18\% over the baseline and three SOTA methods on the polyp dataset. Additionally, it surpassed the second best method (EndoUDA) on the BE dataset by nearly 2\%.

\end{abstract}


\section{Introduction}

The burden and associated costs of GI cancer is increasing rapidly worldwide. In 2020 alone, gastric cancer was responsible for 1.089 million new cases and 0.769 million deaths. This makes this disease the fifth most common type of malignancy and fourth major cause of cancer related mortality \cite{sung2021global}. Although endoscopy is a vital tool in GI cancer screening and surveillance, this technique is still highly operator-dependent and thus 12\% of these types of cancers are often missed on a daily basis \cite{menon2014commonly}.

In order to cope with such issues, AI and in particular computer vision methods have been applied in several Computer Aided Diagnosis (CADx) tools. In the endoscopic surgical domain, image analysis methods have yielded promising outcomes in various downstream tasks such as segmentation, tracking, and detection in clinical settings \cite{ali_comprehensive_2022}. More recently, deep learning-based (DL) based techniques are being increasingly deployed in endoscopic analysis \cite{npj_DM2023} as data becomes more available and procedures become more complex \cite{luo_advanced_2018}.

\begin{figure}[t]
\center
  \includegraphics[width=0.46\textwidth,height=0.25\textwidth ]{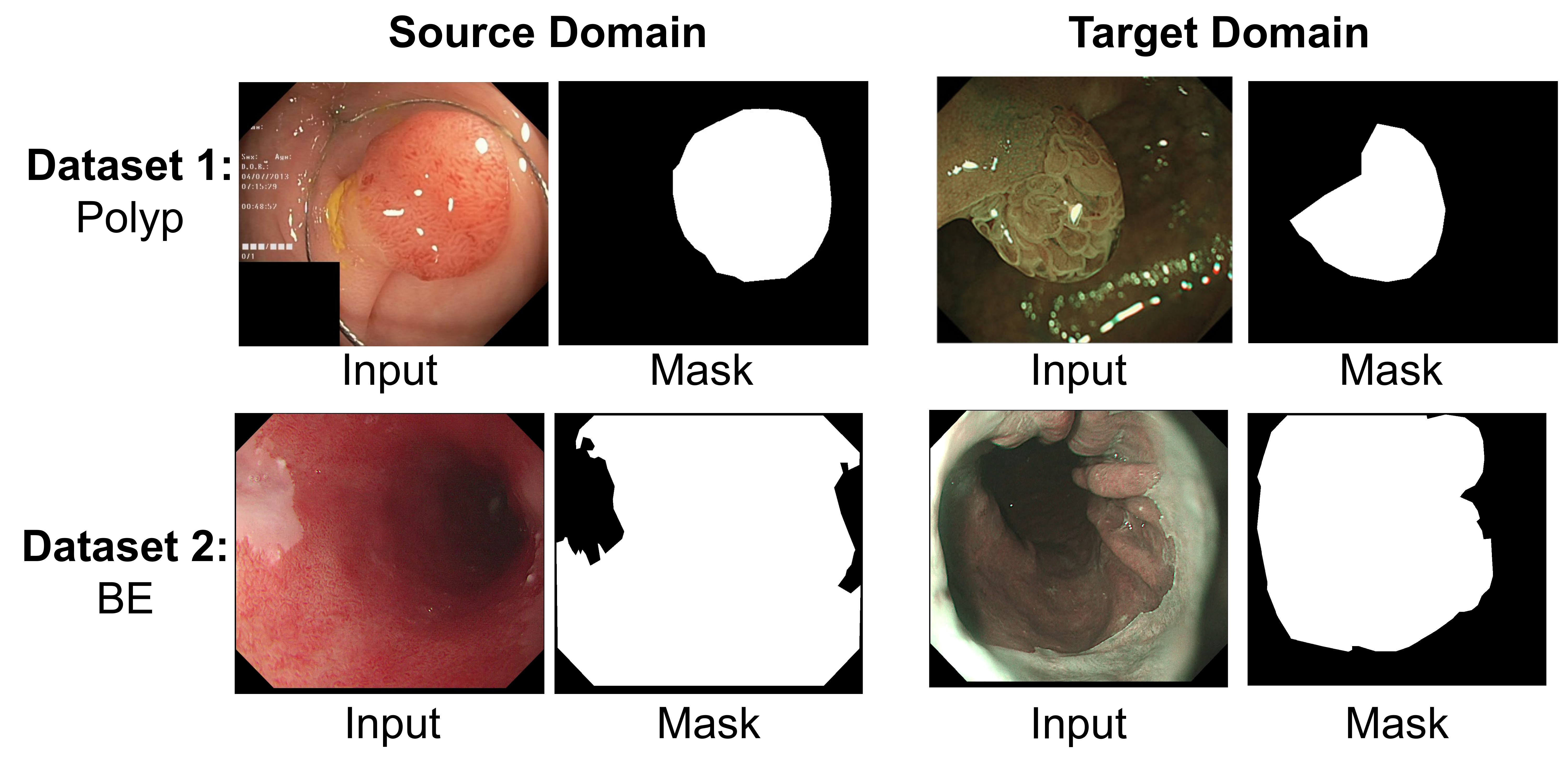}
  \caption{\textbf{Sample images from EndoUDA dataset.} On the left, these show the images acquired with white light imaging (WLI) and on the right, a narrow-band imaging frames (NBI) for polyps and Barret's Esophagus (BE)\cite{celik_endouda_2021}.}
  \label{fig:endouda}
\end{figure}


Despite these recent advances, there are still many limitations associated with DL methods. Traditionally in DL, it is assumed that training and test sets are sampled from the same data distribution. However, when the training set is sampled from a given source domain, whereas the test set is obtained from a similar but different target domain, there will be a domain shift problem. For instance, if a DL model is trained on data from given endoscope of data from a hospital A, it will perform well on the local testing set from the same hospital data. However, the performance of this model  will be sub-optimal if tested on data from another acquisition device (i.e., different vendor) or hospital center B.  In general, sources of distribution shift may occur due to multiple factors, such as the use of different endoscope models, differing imaging modalities, and even different camera parameters. It is therefore essential to develop robust and generalizable models, capable of coping with such data distribution shift problems, while exploiting the existing data as much as possible, as medical data for training is scarce is most situations. 

\section{Motivation and medical context}

In endoscopic imaging, there can be task-specific subdomains where use of specific instrumentation can significantly modify the visual properties of frames being captured, thus producing domain shift problem. More precisely, certain endoscopic examinations for spotting pre-cancerous or cancerous lesions can make use of different lighting modalities such as WLI which is used for a general examination, while  NBI highlights more specific areas. These modalities allow a clinician to inspect different anatomical aspects of the same lesion \cite{barbeiro_narrow-band_2019, kato_magnifying_2010}. This is the case for a disorder called Barrett's Oesophagus (BE), in which the oesophagus contains columnar epithelium rather than the squamous epithelium that is typical for this region of the body. Columnar epithelium is the sort of lining that typically covers the stomach and intestines \cite{spechler_barretts_1986}. The presence of epithelium can enhance the risk of esophageal cancer.   

For a BE segmentation model to work effectively in both imaging modalities (Fig.  \ref{fig:endouda} showing two sample images from the EndoUDA dataset \cite{celik_endouda_2021} using two different modalities),  it is essential that model learns enough discriminant features from the source (WLI) modality to generalize well on the target (NBI) modality as there is a strong indication of change in the visual properties between the two modalities \cite{celik_endouda_2021}. In order to reduce the model complexity and decrease the cost of training, the model should be able to cope with these varying lighting conditions, without requiring modality-specific model training. 

\begin{figure*}[htbp]
\centerline{\includegraphics[width=0.99\textwidth]{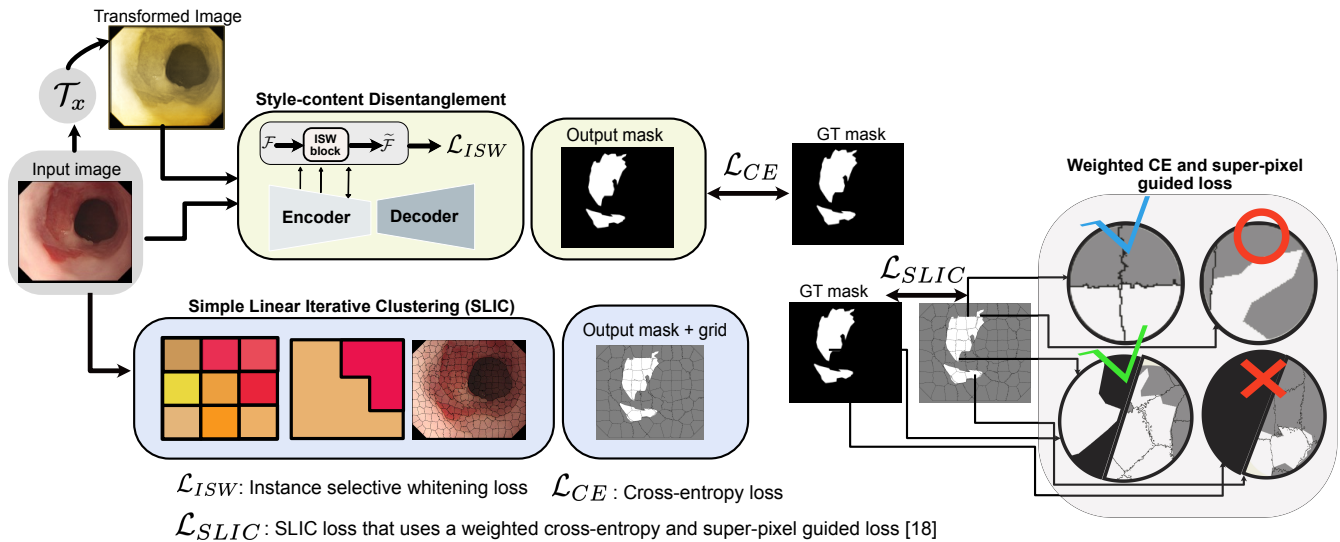}}
\caption{\textbf{Block diagram of our proposed model.} Input image is provided to both segmentation network and Simple Linear Iterative Clustering (SLIC) \cite{achanta_slic_2012} to compute superpixels. We take intermediate feature maps from the ResNet50 backbone to apply ISW transformation to disentangle style-content information. The output prediction mask is then combined with the superpixel grid, where two different loss objectives are computed and combined together denoted as $\mathcal{L}_{SLIC}$: 1) the superpixel guided loss, which assesses how closely the mask follows the superpixel boundaries (a red circle indicates a segmentation that does not follow the object edges, while a blue checkmark indicates a border that is carefully followed). 2) binary cross entroppy loss which determines the overall performance of the network (green checkmark indicates good accuracy while the red cross mark shows poor performance.)}


\label{fig:summary}
\end{figure*}

A number of studies have been conducted to alleviate the domain shift problem in DL models \cite{csurka_unsupervised_2021, tommasi_learning_2016}. These methods can be subdivided into two main groups: i) domain adaptation  (DA) approaches, where models have access to the target domain during training and ii) domain generalization (DG) techniques, where model is  trained solely on source domain and it is designed to perform well on the target domain. There can be several ways to implement these methods such as using regularization or augmentation in the context of training of conventional DL methods; other approaches include exploiting the image style and content information for disentanglement, or using frequency spectrum information\cite{zhou_domain_2022}.

Recently, DG methods have demonstrated a promising performance in semantic segmentation tasks in other areas not related to endoscopy,  such as autonomous driving. Inspired by this previous work, we build upon our own earlier work SUPRA \cite{martinez2023supra}, in which Simple Linear Iterative Clustering (SLIC) was used to generate preliminary segmentation mask with the help of novel SLICloss function to propose an extension  that disentangles style and content information from the image in order to enhance the results obtained by the superpixel consistency method. Thus,  we exploit the feature covariance of original and photometric transformed images to effectively suppress the style information and retain the image content for a better generalization performance of the model, as introduced in \cite{choi2021robustnet}. To evaluate the efficacy of our approach, we use two sub-sets from the EndoUDA dataset, which was introduced to assess DG capabilities in segmentation tasks. The dataset contains images of Barret Esophagus and Polyps in both WLI and NBI imaging modalities. Our experimental results  indicate a significant performance improvement over earlier methods in the literature. 

The rest of this paper is organized as follows: Section III surveys the DG methods developed in the literature. Section IV discusses the proposed DG approach. In Section V we discuss the details of the the experimental setup, with training and testing details. Section VI presents the quantitative and qualitative results of our approach in the EndoUDA dataset. Finally, Section VII presents conclusion and future work. 

\section{State of the art}

DG for semantic segmentation has recently attracted significant attention from the research community. Some earlier approaches have tried to replace batch normalization with instance norm (IN) for better generalization performance. To that extent, IBN-Net \cite{pan2018two} proposed to integrate instance norm and batch norm together in the backbone network to improve model's ability to learn domain-invariant features. Authors in \cite{choi2021robustnet} proposed to use IN while exploiting the feature covariance matrix to suppress domain-specific style. To tackle the limitations of IN, which removes some domain-invariant information along-with the domain specific information, the restitution module was proposed to restore the such task-relevant features  \cite{jin2020style}. Some other approaches using memory banks have been proposed to continuously remember the domain-agnostic knowledge of classes across domains \cite{kim2022pin}.



More recently, yet another approach that has emerged in recent years for DG is based on the definition of soft constraints to improve learned feature representations,  instead of relying on more training data, which is particularly difficult to obtain in the medical field.  For example, authors have used clustering and patch-based constraints to improve DG in segmentation applications in other fields \cite{zhang_transferring_2020}. The main rationale of constraint-based methods is to tackle domain shift problem as an overfit scenario, where constraints have the effect of a regularizer that encourages the model to learn domain-invariant features globally across source domain data. 
In this context, a super-pixel patch consistency constraint was proposed to reduce the domain shift between data of different lighting modalities \cite{martinez2023supra}. 

Superpixel based constraints generate a series of evenly-spaced centers and group pixels based on color similarities and distance from each other achieving very good results in a wide variety of images. The main weakness of such methods is that they rely on domain-speciic hyper-parameter that are difficult to optimize
%
%
To tackle this problem, herein we propose a dual-learning strategy that the super-pixel guided consistency of SUPRA  with feature covariances to suppress domain-specific information for endoscopic segmentation. As our results demonstrate, our proposed approach enables us to produce state of the art results for two imaging modalities, white light imaging and narrow-band imaging in the EndoUDA dataset. 
 
\section{Proposed Approach}

\subsection{Preliminaries:} SUPRA \cite{martinez2023supra} is a supepixel based consistency framework  that was proposed for improved generalization on the EndoUDA BE dataset. The method addressed the DG problem by proposing a loss function to penalize output predictions from a base segmentation model in disagreement with color variations present in the image. The proposed loss was computed by combining Binary cross entropy loss (for improved classification performance) and superpixel guided loss to constrain the network to focus on preserving color consistency. 

Superpixel based consistency can generate evenly spaced centers and can group pixels based on color similarities, which achieved promising outcomes in variety of images. However, the hyperparameter selection and tuning is quite tedious when it comes to superpixels. It is quite difficult to know how many superpixels need to be generated, what weight should be given to any specific color and some spatial boundaries can pose problems and cause large variability in the results. 

Therefore, to alleviate this problem and inspired by the work in \cite{choi2021robustnet}, we augment the SUPRA \cite{martinez2023supra} architecture  with ISW to exploit the feature covariance of the image. We assume that a feature covariance matrix contains both style and content information of image and the variance of original and its photometric transformed image can specifically help us disentangle style-content of the image. The sections below outline the details of  superpixel based consistency and the ISW block.

\subsection{SLIC superpixel generation and SLICLoss}
To alleviate the domain shift problem, SUPRA employs superpixels that can leverage the visual properties which are more globally relevant to the lesions in the dataset. The SLIC superpixel generation algorithm works by using two main parameters \cite{achanta_slic_2012}. The first parameter is $k$, which is the number of superpixels to generate; this parameter enforces the generation of similarly sized regions with spacing $S=\sqrt{N/k}$, where $N$ is the number of pixels in the image. The second parameter of the algorithm is $m$, a constant used to calculate a distance measure used to determine which region a pixel belongs to,

\begin{equation}
D = \sqrt{d_c^2+ \frac{d_s}{S}^2 m^2}
\end{equation}
where $d_c$ is euclidean distance for each color space, and $d_s$ is the euclidean distance between pixels. A higher value of $m$ will encourage compactness, creating regions with a lower area-to-perimeter ratio and more regular shapes. When $m$ is lower, it produces more irregular superpixels that more strictly adhere to areas that present a change in color.
 
\begin{figure*}[t!]
\centerline{\includegraphics[width=0.8\textwidth]{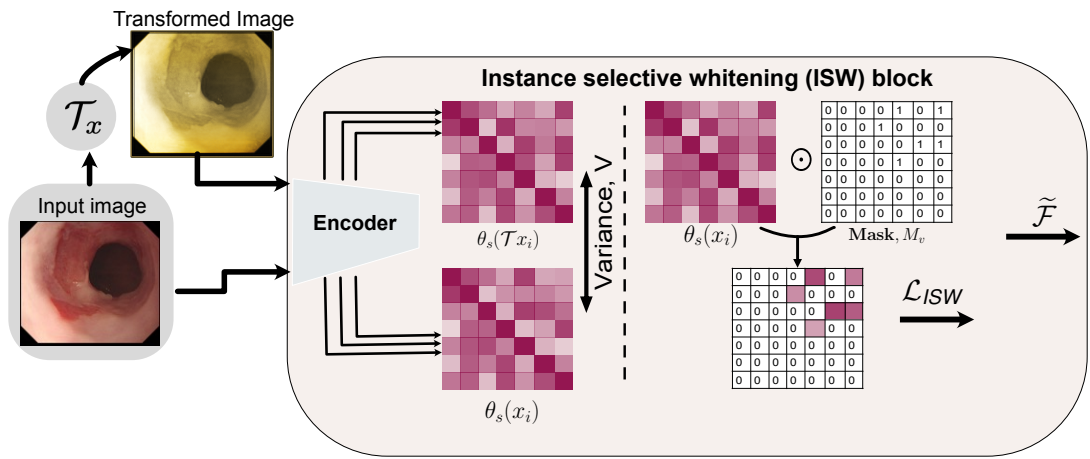}}
\caption{\textbf{Instance selective whitening (ISW) block.} Original input image and its photometric transformed image are passed through the backbone architecture (ResNet50) where feature covariance matrices are computed at three different layers of ResNet50. Variance is computed from both covariance to determine the the mask matrix which is then used to selectively whiten the feature covariance to disentangle style-content information.
}
\label{fig:isw}
\end{figure*}

The loss function is constructed from two main elements as can be seen in Fig. \ref{fig:summary}: The first is Cross Entroppy (CE) represented by $\mathcal{L}_{CE}$ (Fig. \ref{fig:summary}), which evaluates the overall correctness of the prediction mask (y') by comparing it to the ground truth mask (y). The second is a SLIC consistency measure denoted as Superpixel Guided Loss (represented with $\mathcal{L}_{SG}$ in Fig. \ref{fig:summary}). This constraint evaluates whether the results produced are conforming the superpixel segmentation generated by SLIC operating on the input frame ($x_i$).  

The $\mathcal{L}_{SG}$ loss determines the extent to which the superpixel area is occupied by a single class. This is achieved by computing the difference between each class occupied area within a superpixel, and then comparing it with a threshold. Any superpixel that is occupied by less than the threshold is said to be inconsistent with the expected boundary 
(red circle in Fig. \ref{fig:summary}). Every superpixel is evaluated in this manner, with the inconsistencies summed and averaged to produce a final loss metric.

To combine $\mathcal{L}_{CE}$ and $\mathcal{L}_{SG}$, the loss is multiplied by a weighing factor ($\lambda$). The result is then used as the loss for the network.
\color{black}


\begin{equation}
    \mathcal{L}_{SLIC}(x,y,y')= \lambda_1\mathcal{L}_{CE}(y,y')+\lambda_2 \mathcal{L}_{SG}(x,y')
    \label{eq:corr}
\end{equation}

\noindent where $\lambda_1$ and $\lambda_2$ are the weighting factors.  From the eq. \ref{eq:corr}, it can be observed that the weighing factor ($\lambda_2$) can be leveraged to favor more accurate results, or more superpixel consistent results. A higher value of $\lambda$ will boost the effect of the superpixels, but at the cost of decrease in the overall accuracy. The weighting factor ($\lambda_2$), number of superpixel generated ($k$), compactness ($m$), and the threshold for superpixel guided loss ($\mathcal{L}_{SLIC}$) are hyperparameters that must be tuned in accordance to properties in the source domain. For the details on hyperparameter tuning, readers are directed to the earlier work in \cite{martinez2023supra}.

\subsection{Instance Selective Whitening (ISW) block}

Previous literature shows that a whitening transformation (WT) can remove domain-specific style information and boost the overall DG performance \cite{cho2019image,li2017universal,pan2019switchable}. If we denote a feature map as $\mathcal{F}\in \mathbb{R}^{N \times C \times H \times W}$, then a WT is a linear transformation which standardizes features by keeping the variance to 1 while removing the correlations among the channels. The de-correlation process makes the feature covariance matrix ($\theta_s$) close to an identity matrix. 
Earlier approaches to compute WT was using eigen-value decomposition which is highly computationally intensive. Alternatively, the deep whitening transformation (DWT) was proposed in GDWCT \cite{cho2019image} and can be determined by:
\begin{equation}
    \mathcal{L}_{\mathrm{DWT}} = \mathbb{E} [\| \theta_\mu - \mathrm{I} \|_1]
\end{equation}

where $\mathbb{E}$ represents the arithmetic mean. There are two limitations of WT: i) it tends to distort segmentation boundaries \cite{li2017universal} and ii) it reduces feature discrimination \cite{pan2019switchable} because feature covariance matrix ($\theta_s$) contains both domain-specific and domain-invariant information. Therefore, in our framework work we introduce the ISW block to selectively remove the style while retain structural information (see Fig. \ref{fig:isw}). The ISW block takes standardized features (Instance normalization applied to the original features) $\mathcal{F}$ of the original and transformed images from the intermediate layers of the backbone network. The network is initially trained for $n$ epochs ($n$ was empirically set to 5) to obtain stable covariance matrices. Afterwards, the covariance $\theta_s$ for both feature maps, as well as their variance V are computed as follows: 
\begin{equation}
    \theta_s = \frac{1}{h \times w} (\widetilde{\mathcal{F}})(\widetilde{\mathcal{F}})^T
\end{equation}

\begin{figure*}[t]
\centerline{\includegraphics[width=0.6
\textwidth]{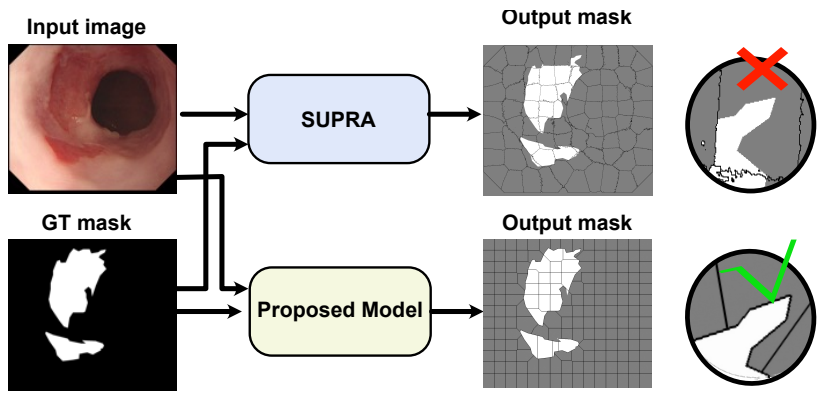}}
\caption{\textbf{Effect of combined loss:} Evaluating the impact of using combined loss of the proposed model and superpixel guided loss. It can be observed in the top part that SUPRA \cite{martinez2023supra} fails to delineate object boundaries which could lead towards poor segmentation output while our proposed architecture does well at correctly segmenting the object of interest. The superpixel grids obtained in the proposed model are at $k$=500.}    
\label{fig:outcome}
\end{figure*}

\begin{equation}
    V = \frac{1}{N} \sum_{i=1}^N \frac{1}{2} ((\theta_s (x_i) - \mu_\theta)^2 + (\theta_s(\mathcal{T}x_i) - \mu_\theta)^2 )
\end{equation} 
\noindent where $\mu_\theta$ represents the mean of both covariance matrices.  It is assumed that high variance values in V indicate the presence of style information which must be suppressed. Therefore, a mechanism to disentangle and separate such values  was implemented using k-means clustering. This results in two distinct groups, $G_{high}$ (containing domain style) and $G_{low}$ (containing content information). Based on this clustered V, we compute the mask $M_v$ and consequently $\mathcal{L}_{ISW}$ as follows, 
\begin{equation}
    \mathcal{L}_{ISW} = \mathbb{E} [| \theta_s \textstyle\bigodot \mathcal{M}_v |]
\end{equation}
The overall cost function for our proposed approach is given by, 

\begin{equation}
    \mathcal{L}_{total} = \mathcal{L}_{task}+  \sum_i^L \lambda \mathcal{L}_{ISW}^i + \mathcal{L}_{SLIC}(x,y,y')
\end{equation}
with $\lambda$ denotes the weight of ISW loss, $L$ represents the number of layers, $\mathcal{L}_{task}$ is cross-entropy loss for semantic segmentation and x, y, y' denote input image, ground truth and prediction respectively. Fig. \ref{fig:outcome} presents the impact of using combination of $\mathcal{L_{ISW}}$ and $\mathcal{L}_{SLIC}$.

\section{Experimental Design}
\subsection{Implementation details}
For the ISW implementation, we used DeepLabv3 as
the baseline model for semantic segmentation with ResNet50 as a backbone. We used SGD as an optimizer and a starting learning rate of 1e-2 followed by polynomial learning rate scheduling with power of 0.9 was used for loss optimization. 
We Incorporated various augmentations such as color jittering, center cropping, Gaussian blur, random cropping, random horizontal flipping and random scaling. The batch size was set as 2 and a center cropping size of 500. 

The model was trained using Microsoft azure 2 NVIDIA Tesla P100-SXM2-16GB GPUs. 
For the superpixel consistency, we resized the images from 
1164 x 1030 to 256 x 256 using bilinear interpolation for all models. We used Geometric data augmentations including horizontal mirroring, rotation, width and height shift, shearing, and zooming with a probability of 5\% except for rotation which was fixed to 20\%.

\subsection{Datasets}
We used EndoUDA dataset which contains endoscopic GI images of two datasets namely, Barrett’s esophagus (BE) and polyps. To evaluate the efficacy of the proposed pipeline we have used clinically acquired white light imaging (WLI)and narrow-band imaging (NBI) modality data for BE and polyps. BE dataset consists of 799 endoscopy images acquired from 68 unique patients of which 515 WLI images are used as source domain data (train set: 80\%, validation set:10\%, and test set: 10\%) and 284 NBI images as target domain data are used for testing. Similarly, for the polyp dataset, we used 1042 images of which 1000 in WLI modality (train set: 80\%, validation set: 10\%, test set: 10\%) and 42 images in NBI modality for testing. 

\begin{table*}[t!]
\centering
\caption{Table showing Intersection over Union (IoU) scores for the baseline, SOTA and our proposed method. Standard deviations ($\sigma$) are provided except SUPRA \cite{martinez2023supra} method. All the results were obtained on models trained on source domain dataset only. The missing values indicate that results were not available for the method. }\label{tab1:results}
\begin{tabular}{lcccc|cccc}
\hline
\textbf{EndoUDA (polyp)}&\multicolumn{4}{c}{\textbf{Test (WLI modality)}}& \multicolumn{4}{c}{\textbf{Target (NBI modality)}}\\
\hline
Method & IoU & Prec. & Rec. & Acc.& IoU & Prec. & Rec. & Acc.\\
\hline
DeepLabv3+(baseline) \cite{chen2018encoder}   & $\mathbf{81.4\pm16.3}$ & 90.0 & 70.0& 89.1& $64.2\pm15.5$ & 80.0 & 80.0 & 78.3 \\
\hline
IBN-Net (CVPR'18) \cite{pan2018two}  & $77.0\pm16.3$ & 80.0 & 90.0 & 85.4 & $68.0\pm15.4$ & 100.0& 70.0 & 75.4  \\
\hline
RobustNet (CVPR'21) \cite{choi2021robustnet}   & $78.1\pm 16.2$ &  80.0 & 90.0 & 85.6 & $70.2\pm 14.8$ & 100.0 & 80.0 & 76.2 \\
\hline
EndoUDA \cite{celik_endouda_2021}   & -- &  -- & -- & -- & $60.5\pm 14.0$ &  72.2 & 70.4 & -- \\
\hline
SUPRA \cite{martinez2023supra}   & -- &  -- & -- & -- & --&  --& --& -- \\
\hline
Ours  & $80.8\pm16.6$ &  90.0 & 90.0 & 87.3 & $\mathbf{78.0\pm}15.0$ & 90.0 & 80.0& 83.8 \\
\hline
\textbf{EndoUDA (BE)}& & \\
\hline
DeepLabv3+(baseline) \cite{chen2018encoder} & $\mathbf{88.2\pm17.2}$ & 90.0 & 90.0 & 91.8& $60.7\pm14.7$ & 70.0 & 90.0 &  65.4\\
\hline
IBN-Net (CVPR'18) \cite{pan2018two}   & $76.8\pm18.6$ & 80.0 & 90.0& 80.5 & $68.3\pm14.2$ & 80.0 & 90.0 & 77.8\\
\hline
RobustNet (CVPR'21) \cite{choi2021robustnet} &  $84.2\pm16.7$ & 90.0 & 90.0 & 87.7 & $71.3\pm13.8$ & 80.0 & 90.0 & 77.9\\
\hline
EndoUDA \cite{celik_endouda_2021}   & -- &  -- & -- & -- & $73.3\pm 14.0$ &  83.2 & 78.4 & -- \\
\hline
SUPRA \cite{martinez2023supra}   & 67.5 &  -- & -- & -- & 64.5&  --& --& -- \\
\hline
Ours & $84.5\pm14.2$ & 90.0 & 90.0 & 87.5& $\mathbf{75.1\pm13.7}$ & 90.0 & 90.0 & 83.2 \\
\hline
\end{tabular}
\end{table*}

\begin{figure*}[htbp]
\centerline{\includegraphics[width=0.7\textwidth]{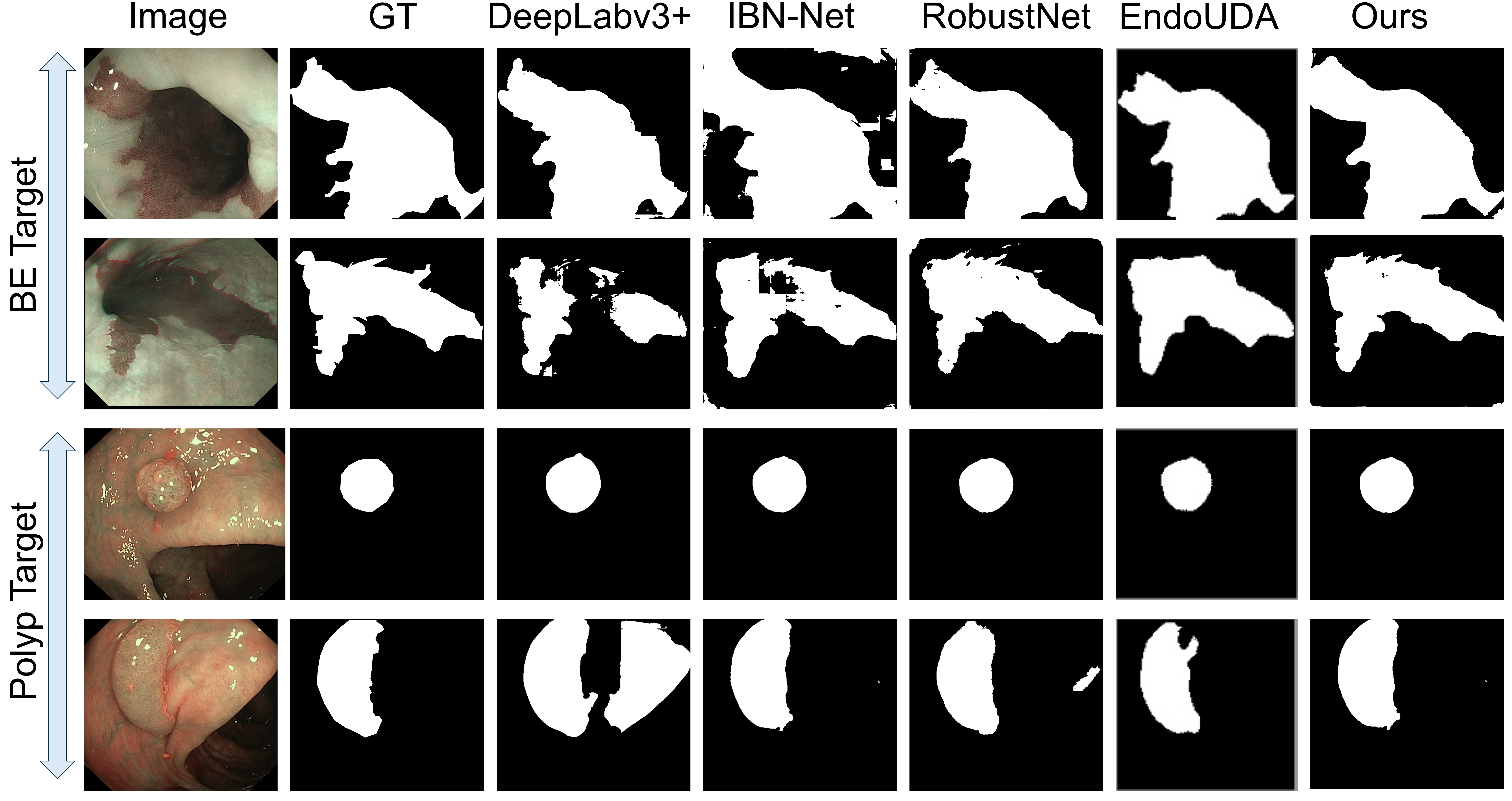}}
\caption{\textbf{Qualitative comparison:} We include frames from tested models on BE and polyp target domain datasets, comparing between DeepLabv3+ \cite{chen2018encoder}, IBN-Net \cite{pan2018two}, RobustNet \cite{choi2021robustnet}, EndoUDA \cite{celik_endouda_2021} and our proposed method. Results indicate that our proposed model performed very close to the ground truth as compared to other methods.}
\label{fig:qual}
\end{figure*}

\section{Results}
In this section, we present the quantitative and qualitative results of baseline models and our proposed approach. We compare our results with three state of the art methods: IBN-Net \cite{pan2018two}, RobustNet \cite{choi2021robustnet}, EndoUDA \cite{celik_endouda_2021} and vanilla baseline method DeepLabv3+ \cite{chen2018encoder}. The results are reported in terms of four different evaluation metrics including intersection-over-union (IoU), precision, recall and accuracy. 

\noindent \textbf{Quantitative results:} We report quantitative results in Table \ref{tab1:results}. It can be observed from Table \ref{tab1:results} that DeepLabv3+ \cite{chen2018encoder} achieved better performance on source domain than our method. However, our approach outperformed all the state of the art methods as well as the baseline on the target domain,  showing better generalizability. This is expected due to the regularization effects provided by SUPRA: some loss in performance is expected in the source domain, with the advantage of enhanced results on unseen data. The results on EndoUDA polyp source domain dataset indicate that DeepLabv3+ \cite{chen2018encoder} achieved 81.4\% IoU score followed by our approach which obtained 80.8\% IoU score. On the target domain, our approach outperformed all state-of-the-art methods and and the Deeplabv3+ \cite{chen2018encoder} baseline model with IoU score of 78.0\%. This is approximately 14\%, 10\%, 8\% and 18\% higher than baseline DeepLabv3+ \cite{chen2018encoder}, SOTA IBN-Net  \cite{pan2018two}, RobustNet \cite{choi2021robustnet}, and EndoUDA \cite{celik_endouda_2021}, respectively.  Similar performance can be observed on EndoUDA  source Barret Esophagus sub-dataset, where DeepLabv3+ \cite{chen2018encoder} achieved the best performance with 88.2\% IoU, whereas our approach obtained second best results with 84.5\% IoU score. On the BE target domain, our approach achieved an IoU score of  71.1\%,  which is substantially higher when compared to the other methods: approximately 15\%, 7\%, 4\%, and 2\% compared to DeepLabv3+ \cite{chen2018encoder}, IBN-Net  \cite{pan2018two}, RobustNet \cite{choi2021robustnet}, and EndoUDA \cite{celik_endouda_2021}, respectively. Similar improvements can
be observed for other evaluation metrics for both datasets and also much higher improvements compared to other state of the art methods.

\begin{table*}[t!]
\begin{center}
\resizebox{0.75\textwidth}{!}{%
\begin{tabular}{l|l|c|c|c}
          & \textbf{Hyperparameter} & \textbf{Source Domain IoU} & \textbf{Target Domain IoU} & \textbf{Fixed Parameters} \\ \hline
$\lambda$ & Weight 50\%             &  67.5                    &       69.8               & $k$ = 100,                \\
          & Weight 75\%* & \textbf{81.6}&\textbf{77.1}       & $m$ = 40                  \\
          & Weight 100\%            &   61.3                    &     55.7                  &                           \\ \hline
$k$       & 50 Superpixels          &    \textbf{78.3}          &       74.9        &  $\lambda$ = 75\%,                        \\
          & 150 Superpixels         &     74.5                &           70.3           &  $m$ = 40       \\
          & 500 Superpixels*        &     76.1                 &      \textbf{78.0}                  &                 \\
          & 1000 Superpixels        &     70.7                &        65.4             &                           \\ \hline
$m$       & 20 Consistency          &     77.8                 &        55.5               & $\lambda$ = 75\%,         \\
          & 30 Consistency& \textbf{79.4}         &    61.8          & $k$ = 100                 \\
          & 50  Consistency*        &    71.9                   &\textbf{74.1}            &                          
\end{tabular}}
\end{center}
\caption{\textbf{Effect of hyperparameters on source and target domains:} Experiments performed to observe the effect of different hyperparameters on our proposed network performance. The validation split from the source domain and target test domain is used to determine the results. We used grid search mechanism to choose the appropriate hyperparameters for the validation spilt from source domain while we used qualitative assessment of the generated superpixel boundaries to select the best values of $\lambda$, $m$, $k$. 
\textbf{$\lambda$:} Weighing factor for the superpixel boundary consistency
$k$: Number of superpixels generated.
$m$: Compactness.  
\textbf{*} Hyperparameter value used for the proposed model. The best results are shown with bold formatting.}
\label{tab2}
\end{table*}

\noindent \textbf{Qualitative results:} We present the qualitative performance of our proposed approach, baseline DeepLabv3+ and state-of-the-art methods in Fig. \ref{fig:qual} on both EndoUDA BE and polyp target domain sub-datasets. It can be observed that our proposed approach produces results more closer to the ground truth than  other methods in comparison. For example, in the case of Barrett's esophagus, the upper regions in the first row is better segmented than the closes EndoUDA. Similarly, for polyp target dataset in the last row our method has improved segmentation compared to the second best methods RobustNet. The performance mostly is degraded significantly on baseline and IBN-Net models. 

\noindent \textbf{Hyperparameter Search}: SLIC-based consistency methods are highly reliant on the hyperparameter values. To that end, we employed two ways to evaluate the effect of hyperparameters on the network performance and to select the best values for three variables i.e., $\lambda$, $m$, and $k$: grid search (on validation source domain data) and qualitative assessment (on target set data). Table \ref{tab2} presents the results for the experiments performed for the hyperparameters search.

\section{Conclusions}
In this work, we proposed a style-content disentanglement pipeline to boost generalization performance with superpixel-based consistency on two domain endoscopic (white light and narrow-band) imaging datasets. Our novel approach improves over the SOTA RobustNet model. As demonstrated by our experiments, it can suppress the domain-specific information, suggesting that the model can learn discriminant features for improved generalization from source to target domain. We evaluated our approach on two gastrointestinal tract (GI) datasets with significant improvements in target modality over all SOTA and the baseline with competitive results on the same modality data. As a future work, we intend to explore different backbone architectures such as UNet and validate the approach on data from different centers.



\bibliography{egbib}{}
\bibliographystyle{IEEEtran}
\vspace{12pt}

\end{document}